\documentclass[10pt,conference]{IEEEtran}
\IEEEoverridecommandlockouts
\usepackage{cite}
\usepackage{amsmath,amssymb,amsfonts}
\usepackage{algorithmic}
\usepackage{textcomp}
\usepackage{textcomp}
\usepackage{multirow, tabularx, balance, stfloats}
\usepackage[numbers]{natbib}
\usepackage{graphicx}
\usepackage[dvipsnames]{xcolor}
\usepackage{url}
\usepackage{comment}

\usepackage{color, colortbl}
\definecolor{Gray}{gray}{0.9}

\def\BibTeX{{\rm B\kern-.05em{\sc i\kern-.025em b}\kern-.08em
    T\kern-.1667em\lower.7ex\hbox{E}\kern-.125emX}}
\begin{document}

\title{Pre-Trained Neural Language Models for Automatic Mobile App User Feedback Answer Generation
\thanks{This research is supported by a fund from Natural Sciences and Engineering Research Council of Canada RGPIN-2019-05175.}
}

\author{\IEEEauthorblockN{Yue Cao}
\IEEEauthorblockA{\textit{Department of Computer Science} \\
\textit{University of British Columbia}\\
Kelowna, Canada \\
caoyuecc@mail.ubc.ca}
\and
\IEEEauthorblockN{Fatemeh H. Fard}
\IEEEauthorblockA{\textit{Department of Computer Science} \\
\textit{University of British Columbia}\\
Kelowna, Canada \\
fatemeh.fard@ubc.ca}

}

\maketitle

\begin{abstract}

  Studies show that developers' answers to the mobile app users' feedbacks on app stores can increase the apps' star rating. 
  To help app developers generate answers that are related to the users' issues,  
  recent studies develop models to generate the answers automatically. 
    \textit{Aims: }
  The app response generation models use deep neural networks and require training data.
  Pre-Trained neural language Models (PTM) used in Natural Language Processing (NLP) take advantage of the information they learned from a large corpora in an unsupervised manner, and can reduce the amount of required training data. 
  In this paper, we evaluate PTMs to generate replies to the mobile app user feedbacks.
    \textit{Method: }
  We train a Transformer model from scratch and fine tune two PTMs 
  to evaluate the generated responses, which are compared to RRGEN, a current app response model. We also evaluate the models with different portions of the training data. 
    \textit{Results: }
  The results on a large dataset evaluated by automatic metrics show that PTMs obtain lower scores than the baselines. 
  However, our human evaluation confirm that PTMs can generate more relevant and meaningful responses to the posted feedbacks. 
  Moreover, the performance of PTMs has less drop compared to other model when the amount of training data is reduced to 1/3. 
    \textit{Conclusion: }
  PTMs are useful in generating responses to app reviews and are more robust models to the amount of training data provided. However, the prediction time is 19X than RRGEN. 
  This study can provide new avenues for research in adapting the PTMs for analyzing mobile app user feedbacks.

\end{abstract}


\section{Introduction}

Mobile applications (apps) have a competitive market with more than $3.14$ million apps on Google Play and $2.09$ million iOS apps as of the forth quarter of 2020 \footnote{https://bit.ly/3EYvxv0}.
Mobile apps are developed in short cycles and multiple short releases prevent testing all features. 
A main resource for users and developers to communicate the app issues and users' interests is app store where users post reviews and developers can respond. 
Previous research show that when developers reply to users' feedbacks, the rating of their app increases \cite{hassan2018studying} by an average of $0.7$ stars by users\footnote{https://bit.ly/3iz7vNJ}, which affects the number of app downloads\footnote{https://bit.ly/2Y6raxi}. 
Therefore, it is important that the app developers respond to the users’ reviews in a timely manner. 
However, replying to each comment manually can be time-consuming. 
Hence, recent studies focus on developing models for replying to the app reviews automatically.
RRGEN is a Recurrent Neural Network (RNN) based model that is trained on a dataset of users' feedbacks and developers' responses from Google Play and is enhanced with app specific information to generate replies \cite{gao2019automatingRRGEN}. 
AARSynth is another RNN-based model which adds information retrieval techniques to the model to help the system generate better replies for different apps \cite{farooq2020app}. 

As training neural networks require a lot of data, Pre-Trained language Models (PTM) are developed, where the model is trained on a large general purpose corpora in an unsupervised manner \cite{devlin2018bert}. 
So, PTM learns a general context about natural language which is later transferred to specific tasks (e.g. question answering, classification) during fine tuning of the model. 
PTMs, therefore, reduce the amount of labeled data required for each task and have shown improvements for many tasks in Natural Language Processing (NLP) \cite{devlin2018bert, liu2019roberta}. 
The main neural network architecture that is used for PTMs is the Transformer architecture \cite{vaswani2017attention}. 
The advantages of using PTMs in software engineering are explored for sentiment classification and code-related tasks (e.g. comment generation) \cite{zhang2020sentiment, hadi2021evaluating, feng2020codebert}.
However, there is no study that evaluates their performance for app review response generation. 
In this paper, we investigate the performance of two PTMs for this purpose. The results using a large dataset from \cite{farooq2020app} are compared against a Transformer model that is trained from scratch and RRGEN, the state of the art app review reply generation. 
In this study, we intend to answer the following research questions. 

\textbf{RQ1) What is the performance of the Transformer-based PTMs in response generation for app reviews? }
Here, we train a Transformer model from scratch and also fine tune the PTMs for response generation. The results are compared to RRGEN, using automatic evaluation metrics as well as the time required to generate the responses. 
The Transformer model will show the effect of the neural architecture. 

\textit{PTMs obtain lower scores using automatic evaluation metrics, and have a longer prediction time. Transformer on the other hand achieves the best score. RNN-based model is the fastest in reply generation.}

\textbf{RQ2) What is the effect of the size of training dataset on the performance of the models?}
Obtaining high quality training data is expensive and not many reviews contain a unique reply. 
In this RQ, we investigate the performance of all models when 1/3, 2/3, or full training dataset is used. 

\textit{Among all, PTMs are more robust when training data is reduced. The Transformer model is less impacted by this reduction compared to RRGEN.}

We also provide the results of human evaluations and their preferred answers generated by each model, which shows responses by PTMs are more accurate, informative, and fluent. 
Our main contribution in this paper is the empirical studies of Transformer model and PTMs for app review reply generation.

The rest of the paper is organized as follows. 
Section \ref{background} provides the required background. 
The details of our experimental set up and results are provided in Sections \ref{experimental-setup} and \ref{results}, followed by human evaluation results in Section \ref{human-eval}. 
Section \ref{discussions} is dedicated to discussions. 
Threats to validity, related works, and conclusion are discussed in Sections \ref{threats}, \ref{related-works}, and \ref{conclusion}. 

\section{Background and Selected Models} \label{background}

\textbf{Transformers. }
Transformer \cite{vaswani2017attention} is the state-of-the-art NLP model that employs self-attention and multi-head attention techniques. The self-attention mechanism allows the model to analyze other words in the input sentence when a word is encoded. 
The multi-head attention 
helps the model pay attention to different positions in the input sentence (expanding the self-attention) and have different representation subspaces for each head, which are then averaged and enable the model to have better performance in sentence understanding.

\textbf{BERT. }
Bidirectional Encoder Representations from Transformers (BERT) \cite{devlin2018bert} is a Transformer-based PTM, which is trained on a large natural language corpora and can be fine tuned on different downstream tasks. 
It uses the Masked Language Model and Next Sentence Prediction as the training tasks. The former predicts the MASKed tokens in the input, and the latter predicts whether a sentence is coming after another sentence. 
BERT has achieved high scores on different NLP tasks. 
As this is the base model for many PTMs, we use BERT in our studies. 
In addition, Rothe et al. \cite{rothe2020leveraging} show the effectiveness of applying the BERT model to a sequence-to-sequence (seq2seq) task such as translation or generation tasks when two pre-trained BERT models are used to build a seq2seq encoder decoder model. This is similar to the task of reply generation which requires a seq2seq model; where the input to the encoder is an app review and the model should generate its response in the decoder part of the model, therefore, translating the input review to an output response. 

\textbf{RoBERTa. }
Liu et al. \cite{liu2019roberta} proposed the Robustly Optimized BERT Pretraining Approach model. This is based on the BERT architecture, but used more training data compared to BERT and it was trained on longer sequences. RoBERTa uses dynamic masking and achieved state of the art performance in various downstream tasks after publication. Therefore, we use RoBERTa in our studies for app review response generation.

\section{Experimental Setup} \label{experimental-setup}

\subsection{Dataset }
Farooq et al. \cite{farooq2020app} released a large dataset containing app reviews and developers' responses. 
This dataset contains 103 unique applications from Google Play and has 47 apps overlapped with RRGEN's dataset \cite{gao2019automatingRRGEN}. As this is a larger dataset than the one used in RRGEN's study, we used this dataset. 
The dataset consists of 570,881 review-response pairs in total. 
However, the responses for some apps in the dataset are identical, meaning that the response is always the same no matter what comments the user has posted.
So, the reply is not pointing to the exact issue that users expressed. 
As we aim to develop a model that generates responses that correspond to each review specifically, we only keep one review-response pair for each template when duplicate responses are found. 
An example of such duplicate reply is: 
\textit{Well that's no good! We're sorry to hear that and want to resolve this issue asap. Please visit our FAQ at URL for troubleshooting tips. Hopefully, you'll be back to swiping in no time!}
This response is found for the following issues users posted for this app:

Review1: \textit{I keep getting notifications of new messages, but there are no new messages. What is the issue now?}

Review2: \textit{Cannot access the application. Already have account also. Please help \textlangle{}signature\textrangle{}.}


Following previous studies \cite{gao2019automatingRRGEN}, we lower case all strings and replace all URLs, numbers, app names, and greetings with their corresponding labels, \textlangle{}URL\textrangle{}, \textlangle{}NUMBER\textrangle{}, \textlangle{}APP\textrangle{}, and \textlangle{}SALUTATION\textrangle{}, respectively. 
Similar to \cite{gao2019automatingRRGEN}, we limit the maximum length of each review or response to 256 tokens. 
After these steps, 239,360 review-response pairs are left in the dataset. 
We randomly split the data into train, validation, and test sets by 8:1:1 portions; leaving 191,511, 23,937, and 23,912 pairs in each one.

\subsection{Training Details }
We train a Transformer model, which is based on the original Transformer proposed by Vaswani et al. \cite{vaswani2017attention}, from scratch. 
The batch size and the hidden dimensions are set to 32 and 512, respectively. 
Stochastic gradient descent (SGD) with the momentum of $0.9$ and the initial learning rate of $0.1$ is used as the optimizer. 
We trained the model for 25 epochs and the best checkpoint was selected according to validation losses, taking care of not overfitting the model.

For both PTMs, we use the HuggingFace library\footnote{\url{https://huggingface.co/}} 
which has the state of the art models. 
As both PTMs have a stack of encoders only, we used the encoder decoder model \cite{rothe2020leveraging} from HuggingFace to connect two BERT models to form a seq2seq model: one as encoder and one as decoder. The same process is used for RoBERTa. 
BERT and RoBERTa have two model types, base and large. 
As the large models require more memory, we choose the base models. 
We use the training and validation sets to fine tune each PTM, without changing the hyperparameters. 
The best checkpoint based on validation losses is chosen. 
All experiments in this study are performed on a cloud server using NVIDIA Tesla V100 GPU.
We train all models (including RRGEN) fully, and use the losses to prevent overfitting.

\subsection{Evaluation Metrics}
We use the automatic metrics, BLEU-n and ROUGE-L, that are used to evaluate the generation tasks. These metrics evaluate the similarity between the ground truth and the generated text, and are used in previous app response studies \cite{gao2019automatingRRGEN, farooq2020app}. 
BLEU-n \cite{papineni2002bleu} uses n-gram (i.e. contiguous sequence of n tokens) precision and a brevity penalty to evaluate the quality of the generated sentence against the ground truth (i.e., developer's response). 
Variable n is the number of contiguous tokens to be considered. 
We use BLEU-(1,2,3,4) scores in the evaluations. 
Unigram is applied to evaluate the accuracy of each single word in the predicted sentence, while BLEU-4 is adopted to measure the fluency of the sentence. 
ROUGE-L \cite{lin2004rouge} is another metric that uses Longest Common Subsequence and F-measure to evaluate the texts' similarity. 

\subsection{Baseline Approaches}
To evaluate the PTMs, we choose two models. 

\textit{Transformer }
is the state of the art neural network architecture, which we train from scratch (no pre-training). 

\textit{RRGEN: }
At the time of this study, RRGEN \cite{gao2019automatingRRGEN} is the only open source model for app review response generation\footnote{We contacted the authors of AARSynth model \cite{farooq2020app}, but were only able to access their dataset (although the model is mentioned to be open sourced). The first author clarified questions about the dataset that we really appreciate. }. 
We chose RRGEN as it is an up-to-date open-sourced model for app review automatic reply generation. Also, the authors reported a BLEU-4 score of $36.17$.
To generate responses, RRGEN uses an encoder-decoder architecture and for each user review, it incorporates extra information: app category, review length, user rating, a set of keywords from Di Sorbo et al.'s work \cite{di2016would}, and the sentiment of the review. 
As we use another dataset\footnote{The dataset from RRGEN \cite{gao2019automatingRRGEN} is not available and we could not obtain the dataset by contacting the authors. Therefore, we only use the dataset from AARSynth study \cite{farooq2020app} which is on the same topic.}, we followed the practice of RRGEN and extracted all the required information attributed for each review in our dataset. 
We extracted the app categories by mining Google Play based on the app names; and used SentiStrength\footnote{\url{http://sentistrength.wlv.ac.uk/}} 
to find the sentiment score of each review, as it was used by RRGEN. 
The set of keywords and the length of the reviews were also used. 
Hence, all the required attributes were provided to RRGEN, when evaluating it. We set the hyperparamters as recommended for RRGEN.

\begin{table}[h!]
    \centering
    \caption{Results of models for response generation. All scores are in percentages. BLEU scores are shown with B.}
    \label{table:RQ1}
    \begin{tabular}{c|c|c|c|c|c}
      \hline
		\textbf{Models} & \textbf{B-1} & \textbf{B-2} & \textbf{B-3} & \textbf{B-4} & \textbf{ROUGE-L} \\
      \hline
            RRGEN & 32.71 & 21.26 & 16.49 & 13.75 & 26.66 \\
            \hline
            Transformer & \textbf{38.18}	& \textbf{25.33} & \textbf{19.53} & \textbf{15.81} & \textbf{30.20} \\
            \hline
            RoBERTa &23.73	& 12.65 & 8.48 & 6.28 & 19.95 \\
            BERT & 28.16	& 17.14 & 12.17 & 8.88 & 23.46 \\
            \hline

    \end{tabular}
\end{table}

\begin{table}[h!]
    \centering
    \caption{The required time for each model to generate replies. The amounts are shown in seconds (second/response).}
    \label{table:RQ1-time}
    \begin{tabular}{ccccc}
      \hline
		\textbf{Models} & \textbf{RRGEN} & \textbf{Transformer} & \textbf{RoBERTa} & \textbf{BERT}  \\
      \hline
            Time & \textbf{0.07} & 0.5 & 1.31 & 1.33  \\
      \hline
    \end{tabular}
\end{table}

\section{Results} \label{results}

\subsection{RQ1. The Performance of the Transformer Based Models}

The results of evalauting models are shown in Table \ref{table:RQ1}.
Among all models, the Transformer model has the highest scores, which is followed by RRGEN.
This model improves the RRGEN's BLEU1-4 and ROUGE-L by 16.17\%, 19.14\%, 18.43\%, 14.9\%, and 13.2\% respectively.
To provide a fair comparison, we followed all the steps and settings recommended by RREGEN when training the mode. The BLEU-4 score we obtained for RRGEN is much lower than the score reported in their paper. 
We attribute this to the size of the training set used here, which is ~30\% smaller than the training set of RRGEN's dataset. 
Also, the average length in our data is more than RRGEN's data. 
Among the two PTMs, BERT achieves better results. 
Although we expected to receive the highest scores for both PTMs, these models achieve lower results. 
When we investigated the results manually, we found that the PTMs produced more specific responses to the given reviews, compared to the Transformer, although they achieve lower scores. Therefore, we conduct human evaluations which we will report in Section \ref{human-eval}, and discuss the reasons of this contradictory result in Section \ref{discussions}.
We also calculate the time required for each model to generate the responses, as reported in Table \ref{table:RQ1-time}.
Here, RRGEN is the most efficient of all, achieving 0.07 seconds to generate a response, followed by Transformer model, and then the two PTMs. 
This result is expected and is related to the neural architecture used by RRGEN. 
The time required by the two PTMs to generate a response is very close. 
Note that all execution times are within an acceptable range.

\subsection{RQ2. Effect of Size of Training Data} 

In app stores, the amount of data that has pairs of review-response is much less than the reviews without a developer's response. For example, Gao et al. report that from 15,963,612 reviews, only 318,973 have received a response from the app developer \cite{gao2019automatingRRGEN}. 
Hence, we are interested to assess the ability of the models in settings where the amount of training data is reduced. 
Here, we compare the models where they are trained using 2/3 (127,674 pairs) and 1/3 (63,837 pairs) of the training set. 
The validation and test sets are not changed, and the same hyperparameters are used. The results are shown in Table \ref{table:RQ2}. 

When the models are trained using 2/3 of the training set, BLEU-n and ROUGE-L scores of the Transformer, RoBERTa and BERT are dropped by less than one score. However, the scores for RRGEN is dropped by at least 2 points for all metrics. 
When we use 1/3 of the training set, the performance drop of the models becomes more obvious. 
With 1/3 of the training set, BERT and RoBERTa have the lowest drop, at most 1 score for BERT, and less than one for RoBERTa. 
The Transformer model scores drop between 1.23 and 1.92 points. 
Reducing the size of training set to 1/3 has the highest impact on RRGEN's performance, which is dropped by 3.1 ROUGE-L score, 3.23 BLEU-4 and 5.56 BLEU-1 scores. 

This result shows that the PTMs are less sensitive to the changes in the size of the data, as they are still learning some knowledge in the pre-training. Therefore, they still achieve a good performance with 1/3 of the training data. 
However, the Transformer is more impacted by this reduction, although its scores still do not drop significantly. We relate this to the multi attention head mechanisms used in the Transformer which learns the contextual knowledge about the text. 
Finally, RRGEN is most impacted by this setting. As expected, this is mainly related to not using a pre-learned knowledge, and its RNN-based neural architecture.

\begin{table}[h!]
    \centering
    \caption{The BLEU-n (shown by B-n) and ROUGE-L scores of each model when trained with different portions of the training set.}
    \label{table:RQ2}
    \begin{tabular}{cccccc}
      \hline
		\textbf{Models} & \textbf{B-1} & \textbf{B-2} & \textbf{B-3} & \textbf{B-4} & \textbf{ROUGE-L} \\
      \hline
      
      \multicolumn{6}{c}{RRGEN} \\
      \hline
            Full dataset &  32.71 & 21.26 & 16.49 & 13.75 & 26.66 \\
            66.6\% dataset & 29.42	& 18.59 & 14.14 & 11.65 & 24.86 \\
            33.3\% dataset & 27.15 & 16.88 & 12.80 & 10.52 & 23.56\\
      \hline
      \multicolumn{6}{c}{Transformer} \\
      \hline
            Full dataset &  38.18	& 25.33 & 19.53 & 15.81 & 30.20 \\
            66.6\% dataset & 37.89	& 24.74 & 18.87 & 15.15 & 29.75 \\
            33.3\% dataset & 36.95	& 23.51 & 17.61 & 13.90 & 28.79 \\
            
    \hline
      \multicolumn{6}{c}{RoBERTa} \\
      \hline
            Full dataset & 23.73	& 12.65 & 8.48 & 6.28 & 19.95 \\
            66.6\% dataset & 23.33	& 11.95 & 7.80 & 5.64 & 19.60 \\
            33.3\% dataset & 23.68	& 12.21 & 8.00 & 5.81 & 19.38 \\
      \hline
      \multicolumn{6}{c}{BERT} \\
      \hline
            Full dataset & 28.16	& 17.14 & 12.17 & 8.88 & 23.46 \\
            66.6\% dataset & 27.55	& 16.34 & 11.39 & 8.11 & 22.84 \\
            33.3\% dataset & 27.50	& 16.13 & 11.12 & 7.79 & 22.80 \\
    \hline

    \end{tabular}
\end{table}

\section{Human Evaluation} \label{human-eval}

We conduct a human evaluation study to identify whether a response generated by the models is addressing the issues mentioned in the given review. 
For this study, we use the approach of previous works \cite{gao2019automatingRRGEN}. We randomly selected 100 reviews from the test set, and split them into 5 groups, with 20 pairs of distinct review-response in each group. Each group of the reviews is evaluated by four individuals. 
We invited 20 volunteers to participate in the survey. 
The individuals were given a user review, and 5 different answers to that comment, i.e., the developer's response, and the texts generated by RRGEN, Transformer, RoBERTa, and BERT. They were asked to rate the responses for each review. 
To prevent biased ratings, in the survey, we did not reveal whether a text was written by the developers, or by a model, and removed the name of all the models as well. 
The participants were divided into 5 groups, each rating 20 reviews and their responses, to ensure that each of samples are scored four times. 

\begin{figure}[t!]
   \includegraphics[width=0.5\textwidth]{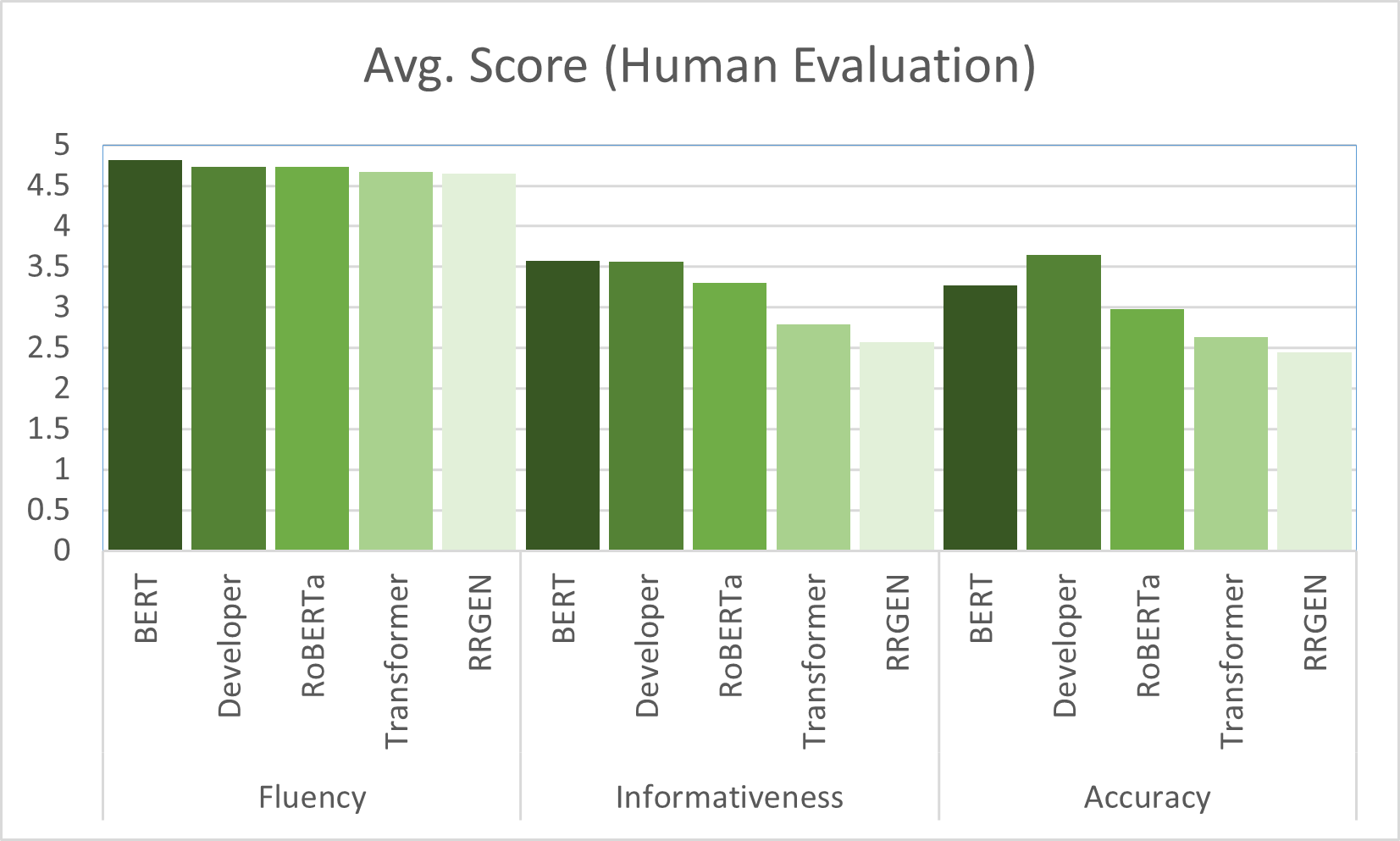}
  \caption{Scores given by individuals to the responses of each model and the developers' responses, separated for each of the three criteria.}
   \label{fig:human-eval1}
\end{figure}

Following previous studies \cite{gao2019automatingRRGEN, farooq2020app, du2018harvesting}, we asked participants to rate the responses based on three criteria: ``grammatical fluency”, ``informativeness” and ``accuracy”. The ``grammatical fluency” measures whether a sentence is semantically smooth, or readable; the ``informativeness” measures the information richness of the sentence; and the ``accuracy” is used to measure whether the response accurately addresses the user’s problem. Users were asked to rate all of these criteria on a scale of 1-5, where 5 is the highest score and 1 is lowest. 
A sample was given to the evaluators prior to ranking, indicating the meaning, examples, and explanation of each criteria.
In total, 400 feedback were collected, and the average scores for each criteria for each of the models are shown in Fig. \ref{fig:human-eval1}. 
Interestingly, BERT has the best score among all models, and even for ``grammatical fluency” and ``informativeness”, BERT's scores are slightly better than the developers' responses. 
For the ``accuracy" criteria, developers' score is 3.64, followed by 3.27 for BERT and 2.98 for RoBERTa. 
Transformer model has lower scores than PTMs for all three criteria, and RRGEN is the lowest among all. 
We believe that the accuracy of the generated responses could be improved by using more attributes in the training (as shown by RRGEN), such as app category and app description. 
Also, adding extra information through information retrieval could improve the accuracy of the produced relies, as explained in \cite{farooq2020app}. 

In addition to these criteria, we asked the evaluators to rank the five replies based on their preferences, where 1 means the most preferred answer and 5 means the less preferred answer. 
In terms of preferences, the developers' responses were the most preferred answers, with an average of 2.3. BERT achieved a score of 2.4, very close to the developers', followed by RoBERTa, Transformer, and RRGEN with scores of 2.7, 3.5, and 4, respectively. 
We discuss these results next. 

\section{Discussions} \label{discussions}

\textbf{Contradictory results of automatic scores and human evaluation for PTMs.}
In RQ1, we reported BLEU-n and ROUGE-L scores for all models, but contrary to our expectations, which come from studies of PTMs for sentiment classification and the NLP studies, the scores of two PTMs were lower than Transformer and RRGEN. 
On the other hand, in the human evaluation, BERT and RoBERTa received better scores than these models. Although the human evaluation is done only on 100 reviews (and can be a possible source of threat in the study as the distribution of this sample may differ from the distribution of the whole dataset), in software engineering it is considered as a reliable resource and is conducted in many studies \cite{gao2019automatingRRGEN, farooq2020app}. 
We associate this contradictory result to the length of the generated sentences. 
Table \ref{table:discussion1} represents the average length of the generated responses for each model, compared to the developers' responses (i.e. ground truth). 
The sentences generated by PTMs are on average longer than other models, and contain approximately 17 and 22 more tokens than the ground truth.
Longer sentences lead to greater variations and deviate from the ground truth. 
BLUE-n and ROUGE-L use the overlap rate of contiguous tokens in generated and target sentences.  
Therefore, if the generated sentences use different grammatical structures or vocabulary, the results of the measurement may not be accurate \cite{sellam2020bleurt}. 
So, although both PTMs are gaining higher human scores and preference, the automatic evaluation metrics are lower. This could be prevented by restricting the PTMs to generate longer sentences, which we avoided in our study, as we want to evaluate the actual capability of the models.

\begin{table}[h!]
    \centering
    \caption{The average reply length of all models.}
    \label{table:discussion1}
    \begin{tabular}{cc}
      \hline
      {Models} & \textbf{Avg. Reply Length} \\
		
      \hline
            Developers & 40.81\\
            \hline
            RRGEN & 42.97\\
            Transformer & 48.87\\
            RoBERTa & 57.47 \\
            BERT & 62.38 \\
    \hline
      
    \end{tabular}
\end{table}

\textbf{The effect of different sizes of training set on the quality of the responses.} 
In RQ2, we used different portions of the training set to train the models. 
Table \ref{table:discussion2} represents a user review complaining about \textcolor{blue}{advertisements} in the app, and the responses by the developers and all models. 
The responses for all models are included when the whole training set, 2/3, or 1/3 of the training data is used. 
The texts in \textcolor{ForestGreen}{green} are the portions that are directly addressing the posted issue. The \textbf{\underline{underlined and bold text}} shows the part that is related, but is not a direct response to the issue.
As it is shown, the generated responses of the PTMs and the Transformer model mention the `ad' problem, and the two PTMs include more information about the ad. 
Here, BERT produces a more accurate response compared to RoBERTa and other models. 
BERT was also evaluated as having better scores and is more preferred by human subjects.
But, the response generated by RRGEN does not mention `ad', and only asks for details about the issue. 

When the training set is reduced to 2/3, the Transformer-based models are still capable of referring to the `ad' problem. 
But, when 1/3 of the training data is used, none of the models mention the `ad' problem in their responses, though the scores of PTMs do not drop significantly. 
This may suggest that the models trained on 1/3 of data still produce grammatically fluent responses, but their accuracy in addressing the users' concern is diluted.

\begin{table*}[h!]
    \centering
    \caption{Samples of generated responses.}
    \label{table:discussion2}
    \begin{tabular}{|p{17.7cm}|}
      \hline
          \textbf{Review:} \textcolor{blue}{too many random ads} \\
		
      \hline
        \textbf{Developer:} \textlangle{}salutation\textrangle{} thanks for supporting \textlangle{}app\textrangle{}, and \textcolor{ForestGreen}{we are sorry for ads}. cm will improve in the future and \textcolor{ForestGreen}{we will try to keep the ads from interfering your use.} if \textcolor{ForestGreen}{you see any ads are inappropriate,} pls let us know. \textlangle{}signature\textrangle{} \\
    \hline
        \rowcolor{Gray}
        \textbf{Whole Training set} \\
    \hline
        \textbf{RRGEN:} \textlangle{}salutation\textrangle{} thanks for using \textlangle{}app\textrangle{}. could you \textbf{\underline{tell us more details about your issue}} via \textlangle{}email\textrangle{} we will do our \textlangle{}signature\textrangle{}\\
        
    \hline
        \textbf{Transformer:} \textlangle{} salutation \textrangle{} we are \textcolor{ForestGreen}{sorry to hear about your concern regarding the ads} . we 're always looking to improve your experience with \textlangle{} app \textrangle{} and we would certainly like to know more about this . send in a quick note to \textlangle{} url \textrangle{} or via the in - app help section and we 'll connect .\\
    \hline
        \textbf{RoBERTa:} \textlangle{}salutation\textrangle{} thanks for using \textlangle{}app\textrangle{} to clean and protect your phone. \textcolor{ForestGreen}{sorry for ads.} could you \textcolor{ForestGreen}{send us some screenshots of the pop up you received} via \textlangle{}email\textrangle{} we will look into the problem and keep optimizing it. if you have any other questions, please feel free to tell us, we are looking forward to your reply. \textlangle{}signature\textrangle{}\\
    \hline
        \textbf{BERT:} \textlangle{} salutation \textrangle{} \textcolor{ForestGreen}{sorry for your inconvenience. about the ads issue,} could yo \textcolor{ForestGreen}{tell us what content is it? in notification bar, home screen, or other places? and the frequency?} could you \textcolor{ForestGreen}{send us some screenshots to show us which kind of ads it is?} \textlangle{} email \textrangle{} look forward to your reply. \textlangle{} signature \textrangle{}\\
        
    \hline
        \rowcolor{Gray}
        \textbf{2/3 Training set} \\
    \hline
        \textbf{RRGEN:} \textlangle{}salutation\textrangle{} \textbf{\underline{sorry for the inconvenience.}} in order to help you solve the problem. \textlangle{}signature\textrangle{}\\
        
    \hline
        \textbf{Transformer:} \textlangle{} salutation \textrangle{} thanks for using \textlangle{} app \textrangle{} . \textcolor{ForestGreen}{sorry for ads} . we 're \textcolor{ForestGreen}{striving to balance user satisfaction and ads} . you can turn charge master off in \textlangle{} app \textrangle{} , tools , charge master . if you have any other questions , please feel free to contact us via \textlangle{} email \textrangle{} \textlangle{} signature \textrangle{}\\
    \hline
        \textbf{RoBERTa:} \textlangle{}salutation\textrangle{} thanks for using \textlangle{}app\textrangle{} to clean and protect your phone. \textcolor{ForestGreen}{sorry for ads.} could you tell us more details about your issue via \textlangle{}email\textrangle{} we will keep optimizing \textlangle{}number\textrangle{} security. if you have any questions or suggestions, please feel free to contact us via feedback(me-setting-feedback). \textlangle{}signature\textrangle{}\\
        
    \hline
        \textbf{BERT:} \textlangle{} salutation \textrangle{} \textcolor{ForestGreen}{sorry for your inconvenience. about the ads issue,} could yo \textcolor{ForestGreen}{tell us what content is it? in notification bar, home screen, or other places? how to happen? and the frequency? what operation caused the problem?} look forward to your \textlangle{} number \textrangle{} - star rate and wish you all the \textlangle{} signature \textrangle{}\\
    \hline
        \rowcolor{Gray}
       \textbf{ 1/3 Training set} \\
    \hline
        \textbf{RRGEN:} \textlangle{}salutation\textrangle{}  you have any question you're welcome to contact \textlangle{}email\textrangle{} . com and it will be helpful for us to solve your problem. \textbf{\underline{could you tell us more details about your issue?}} we will try to improve the app and your  are very important to us. if you have any question or suggestion, please feel free to contact us via \textlangle{}email\textrangle{} \textlangle{}signature\textrangle{}\\
        
    \hline
        \textbf{Transformer:} \textlangle{} salutation \textrangle{} sorry for the inconvenience . \textbf{\underline{could you please tell us which operations on you phone do you think is slow}} you can e - mail me to \textlangle{} email \textrangle{} so we can identify immediately the problem . \textlangle{} signature \textrangle{} \\
    \hline
        \textbf{RoBERTa:} \textlangle{}salutation\textrangle{} thanks for using \textlangle{}app\textrangle{} to clean and protect your phone. \textbf{\underline{could you tell us more details about your issue}} via \textlangle{}email\textrangle{}, me, feedback? we will do our \textlangle{}signature\textrangle{} for your \textlangle{}number\textrangle{} -star rating and wish you a nice day dear!  we appreciate you taking the time to leave us a review.\\
    \hline
        \textbf{BERT:} \textlangle{} salutation \textrangle{} thanks for using \textlangle{} app \textrangle{} to clean and protect your phone. \textbf{\underline{could you tell us more details about your issue}} via \textlangle{} email \textrangle{} we will do our \textlangle{} signature \textrangle{} please feel free to contact us if you have any questions or suggestions. we are always looking for ways to improve our products and we appreciate your \textlangle{} number \textrangle{} stars. \\
    \hline
    \end{tabular}
\end{table*}


\section{Threats to Validity} \label{threats}
\textbf{Internal validity} in our work relates to building the models and replicating the results of RRGEN. For all models, we ensured that the hyperparameters are set properly, and used techniques to prevent overfitting. For RRGEN, we made our best effort to train it with its recommended settings, and provided all the required attributes of the review to train it. 
\textbf{External validity} in our work is related to the generalization of the results and the dataset we used. We used a previously published dataset, and also removed the duplicate records. This dataset is from Google Play and may not reflect the distribution of the reviews and responses from other platforms such as Apple Store.
\textbf{Construct validity} is related to the automatic metrics in our work. We use BLEU-1 to BLEU-4 and ROUGE-L, the frequently used metrics in other software engineering and natural language processing works. We also added human evaluations to assess the quality of the responses for three criteria from human perspective.
\textbf{Conclusion validity} refers to the biases introduced by researchers. In our work, this could be related to human evaluation section. We tried to alleviate this problem by removing the sources of the responses in the questionnaire, and having each question ranked by four people.

\section{Related Works} \label{related-works}

In this section, we briefly explain some of the related works. 
App reviews have been widely studied \cite{martin2016survey} for various purposes such as identifying app issues \cite{palomba2017recommending}, determining the emerging topics of different app versions \cite{gao2019emerging}, and investigating app's privacy \cite{besmer2020investigating}. 
Hassan et al. \cite{hassan2018studying} 
find that responding the reviews are important for app rating. 
Recent studies try to help the developers by generating automatic responses. Gao et al. develop RRGEN model to generate the app review responses, using extra information about apps 
\cite{gao2019automatingRRGEN}. 
Farooq et al. develop AARSynth to generate responses that are aware of the app itself 
\cite{farooq2020app}.
Gao et al. develop CoRe, a response generation system that uses apps' description and information retrieval techniques \cite{CoRe}. Among these three, only RRGEN model is available.
There are a few studies in software engineering that evaluate the capability of PTMs for sentiment analysis \cite{zhang2020sentiment}, user feedback analysis \cite{hadi2021evaluating}, and programming and natural language tasks \cite{feng2020codebert}. 
Our work is different with these studies as we are the first to investigate the application of pre-trained language models and Transformers for app review response generation. 

\section{Conclusion and Future Work} \label{conclusion}
We studied the applicability of PTMs for responding app reviews.
The human subjects confirm that the PTMs produce more accurate, informative, and fluent responses; although the automatic scores are contradictory. We present insights about this difference, which is mainly due to generating longer responses. Additionally, PTMs still show a good performance when the training data is reduced to 33\%, and the Transformer architecture is able to produce better results compared to RNN-based models. 
Future research directions are incorporating the app specific information, using other generation models such as Plug and Play to control the topics, and use lighter Transformer architectures to reduce the computational time.

\bibliographystyle{IEEEtranN}
{\footnotesize\bibliography{references}}

\end{document}